\documentclass[runningheads]{llncs}

\usepackage[T1]{fontenc}
\usepackage{graphicx}
\usepackage{booktabs}
\usepackage{array}
\usepackage{multirow}
\usepackage{amsmath}
\usepackage{amssymb}
\usepackage{algorithm}
\usepackage{algpseudocode}
\usepackage{siunitx}
\usepackage{xcolor}
\usepackage{marvosym}
\usepackage{hyperref}
\hypersetup{
  colorlinks=false,
  hidelinks
}
\usepackage{orcidlink}
\usepackage[capitalize]{cleveref}

\begin{document}
\raggedbottom

\title{Pre- and Post-Treatment Brain Metastases Segmentation Using nnU-Net with Post-Processing for BraTS 2026}
\titlerunning{Brain Metastases Segmentation Using nnU-Net with Post-Processing}

\author{Haobin Liu\inst{1}\,\orcidlink{0009-0008-2174-8217} \and
        Xin Wang\inst{1,2}\textsuperscript{(\Letter)}}
\authorrunning{H. Liu and X. Wang}

\institute{
  College of Software, Jilin University,
  Changchun, China \and
  Key Laboratory of Symbolic Computation and Knowledge Engineering of Ministry of
  Education, Jilin University, Changchun, China \\
  \email{w\_x@jlu.edu.cn}
}

\maketitle

\begin{abstract}
Brain metastases exhibit high inter-lesion variability in size, enhancement
pattern, and post-treatment appearance, making volumetric segmentation of both
pre- and post-treatment cases the central challenge of the BraTS 2026 Task~1
(Brain Metastases). We build a pragmatic pipeline on a 5-fold nnU-Net ResEnc-L
ensemble, in which each fold is trained independently for 1{,}000 epochs with the
standard Dice + cross-entropy loss on 1{,}296 four-modality training cases. This
ensemble is followed by a rule-based post-processing cascade tuned for
the lesion-wise Dice similarity coefficient (LW-DSC), a detection-oriented metric
that behaves very differently from the traditional global Dice. The final
pipeline reaches an LW-DSC of \textbf{0.733 / 0.751 / 0.713 / 0.549} on the
enhancing tumour (ET), tumour core (TC), whole tumour (WT), and resection cavity
(RC) sub-regions on the official validation leaderboard. Rather than trusting these
leaderboard gains, we audit every post-processing stage with a five-fold
out-of-fold (OOF) analysis with no model-training leakage over all $1{,}296$ training cases, scored
with the official BraTS evaluation code (\texttt{BraTS\_evaluation}): it confirms
two stages as robust, per-fold-consistent improvements while the third improves
only the leaderboard and does not reproduce out-of-fold. We further
provide a mechanistic analysis of the LW-DSC metric that explains why
recall-recovering post-processing carries low risk whereas component deletion
does not, and we report thirteen negative results spanning loss engineering,
alternative backbones, and inference-time settings, several of which run counter
to widely held intuitions.
Source code is released under Apache-2.0 at
\url{https://github.com/hornbeamliu/brats2026-met}.

\keywords{Brain metastases segmentation \and nnU-Net \and Post-processing \and
Lesion-wise Dice \and BraTS 2026}
\end{abstract}

\section{Introduction}
\label{sec:intro}

Brain metastases are the most common intracranial neoplasms in adults, and
accurate volumetric assessment underpins radiosurgery planning and
treatment-response monitoring, where manual RANO-BM measurements under-detect
small lesions~\cite{lamba2021epidemiology,ostrom2018epidemiology,%
wang2023stratified,andrearczyk2024automatic}. Within the
long-running BraTS benchmark~\cite{menze2015multimodal,bakas2017advancing,%
bakas2017gbm,bakas2017lgg,baid2021rsna,deverdier2024brats2024,%
karargyris2023federated}, the metastases sub-challenge
(BraTS-MET)~\cite{moawad2023bratsmets,maleki2025analysis} targets the pre- and
post-treatment setting~\cite{bancerek2025nnunets} and, in its 2026 edition, poses
three difficulties that separate it from prior glioma tasks: (i)~cases aggregated
from eight institutions, mixing SRI24-registered and native-space scans, produce
strong inter-site heterogeneity; (ii)~a resection-cavity (RC) label that is
extremely rare, present in only about one in eight training cases, creates
severe class imbalance; and (iii)~the ranking metric is a \emph{lesion-wise} Dice (LW-DSC)
that averages Dice per connected component, rather than the conventional
global-volume Dice, and is therefore dominated by the detection of small
($<\SI{27}{\milli\meter\cubed}$) lesions.

In this short paper we describe our BraTS 2026 Task~1 pipeline (team
\texttt{bin\_JLU}). We keep the architecture and training deliberately close to
the strongest publicly documented nnU-Net baseline for BraTS
glioma~\cite{isensee2024nnunetrevisited,isensee2021nnunet,%
ronneberger2015unet,cicek20163dunet} and concentrate our engineering effort on
inference-time post-processing designed for the LW-DSC metric. The pipeline
couples a 5-fold nnU-Net ResEnc-L ensemble with a rule-based cascade whose two
robust stages are a rule-based clean-up (\texttt{RuleClean}) and a
resection-cavity boundary expansion (\texttt{BoundaryExpand}); a third,
outside-brain false-positive suppression (\texttt{brainF70}), was part of the
submitted Docker container but does not survive our out-of-fold audit and is
reported as a negative result (\Cref{sec:ablation,sec:failed}). Each stage is a
separate script acting on the ensembled volume, so its LW-DSC contribution can
be measured in isolation. On the official Synapse validation leaderboard the
pipeline places all three tumour-side sub-regions (ET, TC, WT) in the leading
tier (\Cref{sec:results}); this is an evaluation on the \emph{validation} set,
not a hidden-test or final-ranking result.

Rather than relying on the leaderboard alone, we audit each stage with a
five-fold out-of-fold analysis over all $1{,}296$ training cases with no
model-training leakage (\Cref{sec:ablation}): this confirms \texttt{RuleClean}
and \texttt{BoundaryExpand} as robust across folds, whereas the \texttt{brainF70}
gain appears only on the leaderboard and falls within out-of-fold variance.
Finally, we document thirteen unsuccessful attempts (\Cref{sec:failed}) across
loss engineering, alternative backbones, and inference-time settings; several run
counter to common intuition: notably, finer sliding-window overlap and
imbalance-oriented losses can each degrade performance, so future participants
need not revisit them.

\section{Methods}
\label{sec:methods}

\subsection{Dataset}
\label{sec:data}

We use the BraTS-MET 2025 training dataset~\cite{maleki2025analysis} as
released for the 2026 challenge: pre- and post-treatment multi-parametric MRI of
brain metastases aggregated from eight institutions, mixing
SRI24~\cite{rohlfing2010sri24}-registered and native-space acquisitions. Each
case provides four modalities --- post-contrast T1-weighted (T1c), native
T1-weighted (T1n), T2-weighted FLAIR (T2-FLAIR), and T2-weighted (T2w) --- and a
5-class annotation: background (0), non-enhancing tumour core (NETC, 1),
surrounding non-enhancing FLAIR hyperintensity (SNFH, 2), enhancing tumour
(ET, 3), and resection cavity (RC, 4). The evaluation regions are
\mbox{TC = NETC $\cup$ ET} and \mbox{WT = NETC $\cup$ SNFH $\cup$ ET}.

The 2026 release provides all four modalities for every one of the
\textbf{1{,}296} training cases, encoded in the nnU-Net~\cite{isensee2021nnunet}
four-channel format (\texttt{\_0000}--\texttt{\_0003} $=$ T1c, T1n, T2-FLAIR,
T2w). Every training case therefore carries a native T2w volume, and we perform
no T2w synthesis or zero-imputation.

Validation is done on the official 179-case Synapse validation
set, in which every case likewise provides all four modalities including a
native T2w. Both the training and the official validation data used in this
work are therefore complete-modality; the inference-time fallback described in
\Cref{sec:inference} is a safeguard for the hidden test phase and is never
exercised on the data reported here. All local out-of-fold (OOF) ablations reported in
\Cref{sec:ablation} are computed on the 1{,}296 training cases using
each case's held-out fold's prediction, and are therefore independent of
the Synapse validation submission budget.

\subsection{Backbone architecture}
\label{sec:backbone}

We adopt the residual encoder variant of nnU-Net (ResEnc-L) introduced in
the ``nnU-Net Revisited'' study~\cite{isensee2024nnunetrevisited}. Its
residual blocks follow the design of ResNet~\cite{he2016resnet}, and its feature
maps are normalised with instance normalisation~\cite{ulyanov2016instancenorm}
as is standard in nnU-Net. We use the
official \texttt{nnUNetPlannerResEncL} planner and the resulting
\texttt{nnUNetResEncUNetLPlans} configuration. All network hyperparameters,
including patch size, batch size, number of pooling stages and per-stage
channel widths, are set automatically by the planner from the dataset
fingerprint and are not manually tuned. We use the \texttt{3d\_fullres}
configuration exclusively; no cascade or low-resolution stage is trained.
The final architecture consists of a residual encoder-decoder with deep
supervision at all resolutions, four input channels, and five output
channels corresponding to background plus the four foreground classes.

\subsection{Training}
\label{sec:training}

Each of the five folds is trained \emph{independently} with the standard
nnU-Net trainer \texttt{nnUNetTrainer\_ResEncL} for a full 1{,}000 epochs.
We use SGD with Nesterov momentum $\mu=0.99$ and weight decay
$3 \times 10^{-5}$, and the standard nnU-Net polynomial learning-rate schedule
with initial learning rate $\eta_0 = 0.01$ and power $0.9$; adaptive
optimisers such as Adam~\cite{kingma2015adam} and AdamW~\cite{loshchilov2019adamw}
are known to underperform SGD on nnU-Net's compound loss and were not
considered here. The optimisation
objective is the standard nnU-Net compound loss $\mathcal{L}=\mathcal{L}_{\text{Dice}}+\mathcal{L}_{\text{CE}}$,
where the Dice term follows the formulation popularised by
V-Net~\cite{milletari2016vnet},
with deep supervision at every decoder scale, without any loss re-weighting or
focal / top-$k$ modification (see \Cref{sec:failed} for negative results on
these alternatives). Data
augmentation follows the nnU-Net default recipe (random mirroring, elastic
deformation, rotation, scaling, additive Gaussian noise, Gaussian blur,
multiplicative brightness / contrast, low-resolution simulation, and gamma
correction). Training is performed with automatic mixed precision (fp16).
We train each fold on 2$\times$NVIDIA RTX A6000 (48\,GB VRAM) using the standard
$\text{batch}=2$ split that the nnU-Net planner selects for
\texttt{3d\_fullres}.

\subsection{Post-processing cascade}
\label{sec:postproc}

The cascade is executed as three independent scripts run in sequence on the
5-fold ensemble prediction. We deliberately keep them as separate stages
rather than fuse into a single entry point, so that each stage's effect on
the LW-DSC can be quantified separately (\Cref{sec:ablation}).

\subsubsection{Stage 1: rule-based clean-up (\texttt{RuleClean}).}
The clean-up stage applies four rules to the argmaxed segmentation volume, in the
order and with the exact semantics specified in \Cref{alg:ruleclean}. All four
operate on 26-connected components rather than isolated voxels, and every removed
voxel is relabelled to background.
\texttt{ET-min10} and \texttt{RC-min10}
remove connected components of ET and RC whose physical
volume is below $10\,\text{mm}^{3}$, respectively; the $10\,\text{mm}^{3}$
threshold was selected because it is well below the challenge's
$27\,\text{mm}^{3}$ small-lesion evaluation cut-off yet still removes the
long tail of single-voxel spurious predictions. \texttt{SNFH-adj2mm} removes an
entire SNFH connected component unless at least one of its voxels lies within
$2\,\text{mm}$ of a TC voxel; because retention is all-or-nothing per component,
the interior of a retained edema component is never eroded. This mirrors the
clinical observation that peritumoural FLAIR hyperintensity should be attached to
the tumour body. \texttt{NETC-adj2mm} applies the same whole-component rule to
NETC with ET as the anchor, enforcing the BraTS-MET annotation guideline
that NETC ``must be enclosed by enhancing tumour''; the $2\,\text{mm}$
tolerance absorbs slight boundary imprecision at the ET/NETC interface. Applied in isolation to the raw 5-fold ensemble, \texttt{RuleClean}
reaches an LW-DSC average across ET / TC / WT / RC of \textbf{0.6823} on the Synapse
validation set, a \textbf{$+0.0305$} gain over the un-post-processed ensemble
baseline.

\begin{algorithm}[t]
\caption{\texttt{RuleClean} (Stage~1). All operations act on 26-connected
components; every removed voxel is relabelled to background. Labels:
$1{=}$NETC, $2{=}$SNFH, $3{=}$ET, $4{=}$RC; the tumour core is
$\text{TC}=\{1,3,4\}$.}
\label{alg:ruleclean}
\begin{algorithmic}[1]
\Require argmax segmentation $S$; voxel spacing $(s_x,s_y,s_z)$ in mm
\Function{RemoveSmall}{$S,\ell,\tau$}
  \State $\tau_{\text{vox}} \gets \max\!\big(1,\ \mathrm{round}(\tau / (s_x s_y s_z))\big)$
  \ForAll{26-connected components $C$ of $\{v : S[v]=\ell\}$}
    \If{$|C| < \tau_{\text{vox}}$} \State $S[C] \gets 0$ \Comment{delete whole component}
    \EndIf
  \EndFor
  \State \Return $S$
\EndFunction
\Statex
\Function{KeepIfTouch}{$S,\ell,A,d$} \Comment{$A$: anchor mask}
  \State $r \gets \max\!\big(1,\ \mathrm{round}(d / \min(s_x,s_y,s_z))\big)$
  \ForAll{26-connected components $C$ of $\{v : S[v]=\ell\}$}
    \If{$\mathrm{dilate}(C, r) \cap A = \varnothing$} \State $S[C] \gets 0$
    \EndIf
  \EndFor
  \State \Return $S$
\EndFunction
\Statex
\State $S \gets \Call{RemoveSmall}{S,\ 3,\ 10\,\text{mm}^{3}}$ \Comment{\texttt{ET-min10}}
\State $S \gets \Call{RemoveSmall}{S,\ 4,\ 10\,\text{mm}^{3}}$ \Comment{\texttt{RC-min10}}
\State $S \gets \Call{KeepIfTouch}{S,\ 2,\ \{v : S[v]\in\text{TC}\},\ 2\,\text{mm}}$ \Comment{\texttt{SNFH-adj2mm}}
\State $S \gets \Call{KeepIfTouch}{S,\ 1,\ \{v : S[v]=3\},\ 2\,\text{mm}}$ \Comment{\texttt{NETC-adj2mm}}
\State \Return $S$
\end{algorithmic}
\end{algorithm}

\subsubsection{Stage 2: resection-cavity boundary expansion (\texttt{BoundaryExpand}).}
This stage addresses a specific, systematic failure of the trained ensemble: it
\emph{under-segments} the resection cavity, predicting RC components whose
contours are truncated inside the true cavity (raw RC lesion-wise DSC is only
$0.367$). LW-DSC on the RC channel is particularly sensitive to this boundary
shrinkage, because a case with a small predicted RC component that is a subset
of a larger ground-truth RC is penalised much more strongly under lesion-wise
matching than under global Dice. We therefore grow each predicted RC component
outward by a fixed margin in that case's own native voxel grid, the same
space in which the network makes its prediction, using a
background-restricted 3D ring dilation. The operation is thus motivated by the
observed under-segmentation rather than by any target physical distance, and it
is region-protected (never entering ET/NETC/SNFH). Let $R_0$ denote
the set of voxels labelled RC after Stage 1, and let $P_{\text{RC}}$ denote
the ensembled RC softmax probability. Starting from $R^{(0)} = R_0$, we
iterate for $i = 1, 2, 3$:
\begin{equation}
R^{(i)} = R^{(i-1)} \;\cup\; \Big\{ v \in \partial R^{(i-1)} \;\Big|\;
   \text{label}(v) = \text{background} \;\wedge\;
   P_{\text{RC}}(v) > 0.10 \Big\},
\end{equation}
where $\partial R^{(i-1)}$ is the 26-connected boundary shell of $R^{(i-1)}$.
The gate on $P_{\text{RC}}$ is a real, active safeguard in the submitted
pipeline: a background voxel is admitted into RC only where the ensembled RC
probability exceeds $0.10$, so the expansion can never grow into regions the
network considers confidently non-RC. It is deliberately set as a permissive
lower bound rather than a tight threshold. The observation that replacing
$0.10$ by $0$ yields near-identical outputs is the expected behaviour of such a
lower bound: voxels immediately adjacent to a predicted RC wall almost always
already carry $P_{\text{RC}}>0.10$, so the gate binds only where the model is
\emph{not} confident (precisely the cases it is meant to block) and is
otherwise rarely called upon. Together with the region protection (Stage~2
never enters ET/NETC/SNFH) and a per-component cap that limits growth to at most
$3\times$ the original component volume, the gate is one of three independent
constraints that keep the operator conservative rather than a blind dilation.
More aggressive settings ($i=5$ dilations with no gate) are catastrophic in our
setting; see \Cref{sec:failed}. In the five-fold out-of-fold audit of
\Cref{sec:ablation}, this stage produces a small but sign-consistent RC gain on
every fold and leaves the ET\slash TC\slash WT channels untouched, confirming it as a safe,
if modest, operator.

\subsubsection{Stage 3: outside-brain false-positive suppression
(\texttt{brainF70}).}
A third stage targets RC components placed partly outside the brain. Exploiting
the skull-stripped T1c (intensity $0$ outside the parenchyma), it defines a
brain mask $B=\{v:\text{T1c}(v)>0\}$ and deletes any RC component whose brain
fraction $|C\cap B|/|C|$ is below $0.70$; no brain-extraction model is needed.
It was part of the submitted Docker container but does not survive the five-fold
out-of-fold audit (\Cref{sec:ablation}). We therefore exclude it from the
recommended cascade, which is \texttt{RuleClean} followed by
\texttt{BoundaryExpand}, and report it as a negative result (\Cref{sec:failed}).

\subsection{Inference and ensembling}
\label{sec:inference}

Inference uses the standard nnU-Net sliding-window predictor with Gaussian
importance weighting at the default step size of $0.5$ (50\% overlap); a sweep
at $0.25$ with three ensemble weightings was systematically worse
(\Cref{sec:failed}). The five per-fold softmax volumes are fused by arithmetic
mean and decoded with \texttt{argmax}; weighted averaging gave no improvement.
Test-time mirroring along all three axes is enabled, as in the nnU-Net default.
The Task~1 environment provides a single $24\,\text{GB}$ A10G GPU, so the
container loads the five fold models one at a time, accumulating softmax on disk
before fusion; this is the configuration behind the \Cref{sec:main} results.

For the hidden test phase the container defines a deterministic missing-modality
fallback: a case lacking native T2w reuses its T2-FLAIR for the
\texttt{\_0003} channel (the closest T2-weighted contrast) rather than
zero-filling. All cases reported here provide native T2w (\Cref{sec:data}), so
this fallback is never exercised.

\section{Results}
\label{sec:results}

We report lesion-wise Dice similarity coefficient (LW-DSC) for each of the
four evaluation sub-regions (ET, TC, WT, RC), following the challenge
protocol \cite{maleki2025analysis,maier2024metrics,reinke2024understanding}.
The primary summary statistic is the arithmetic mean of the four per-region
LW-DSC values (denoted \texttt{avg4}). We complement the primary metric with
per-region small-instance $F_1$ where relevant.

\subsection{Main validation results}
\label{sec:main}

\Cref{tab:main} summarises the score of our final submission on the official
Synapse validation set (179 cases). This submission is the configuration behind
our Docker container: the 5-fold ResEnc-L ensemble described in
\Cref{sec:inference} followed by the post-processing cascade of
\Cref{sec:postproc}. It includes the \texttt{brainF70} stage, which we retain
here only to reproduce the exact submitted leaderboard score; as shown by the
out-of-fold audit in \Cref{sec:ablation}, \texttt{brainF70} is not a reliable
component, and our recommended cascade is \texttt{RuleClean} followed by
\texttt{BoundaryExpand}.

\begin{table}[t]
\centering
\small
\caption{Main validation results on the official Synapse validation set
(179 cases); ``Final pipeline'' is the submitted Docker configuration and avg4
is in bold. See the text for the \texttt{brainF70} caveat.}
\label{tab:main}
\begin{tabular}{@{}lccccc@{}}
\toprule
Submission     & ET     & TC     & WT     & RC     & avg4 \\ \midrule
Final pipeline & 0.7328 & 0.7510 & 0.7131 & 0.5487 & \textbf{0.6864} \\ \bottomrule
\end{tabular}
\end{table}

The resection cavity remains the sole critical deficit relative to our own
tumour-side numbers; we discuss the gap in \Cref{sec:discussion}.

\subsection{Ablation of the post-processing cascade}
\label{sec:ablation}

Consuming the Synapse validation submission budget for a full ablation of
the three post-processing stages would have been prohibitive. We therefore
evaluate the cascade on the complete five-fold OOF predictions
produced by the nnU-Net trainer: for each fold we score its held-out validation
split with that fold's own single model, so every one of the $1{,}296$ training
cases is scored exactly once by a model that never saw it. This gives an
estimate with no model-training leakage on two orders of magnitude more cases than a single
Synapse submission would allow. Scoring uses the official BraTS evaluation code
(\texttt{BraTS\_evaluation}), which wraps the Panoptica lesion-wise evaluator in
its \texttt{mets} configuration (lesion-volume threshold $27\,\text{mm}^{3}$,
overlap $0.2$); regions absent from the ground truth are skipped rather than
scored, exactly as on the leaderboard. The resulting OOF
absolute values are not directly comparable to the ensemble leaderboard numbers
of \Cref{sec:main} (they come from single models and a different case
population), so we read \Cref{tab:ablation} only for the \emph{sign and
per-fold consistency} of each stage's increment, never against the leaderboard
scale.

\begin{figure}[!t]
\centering
\includegraphics[width=\linewidth]{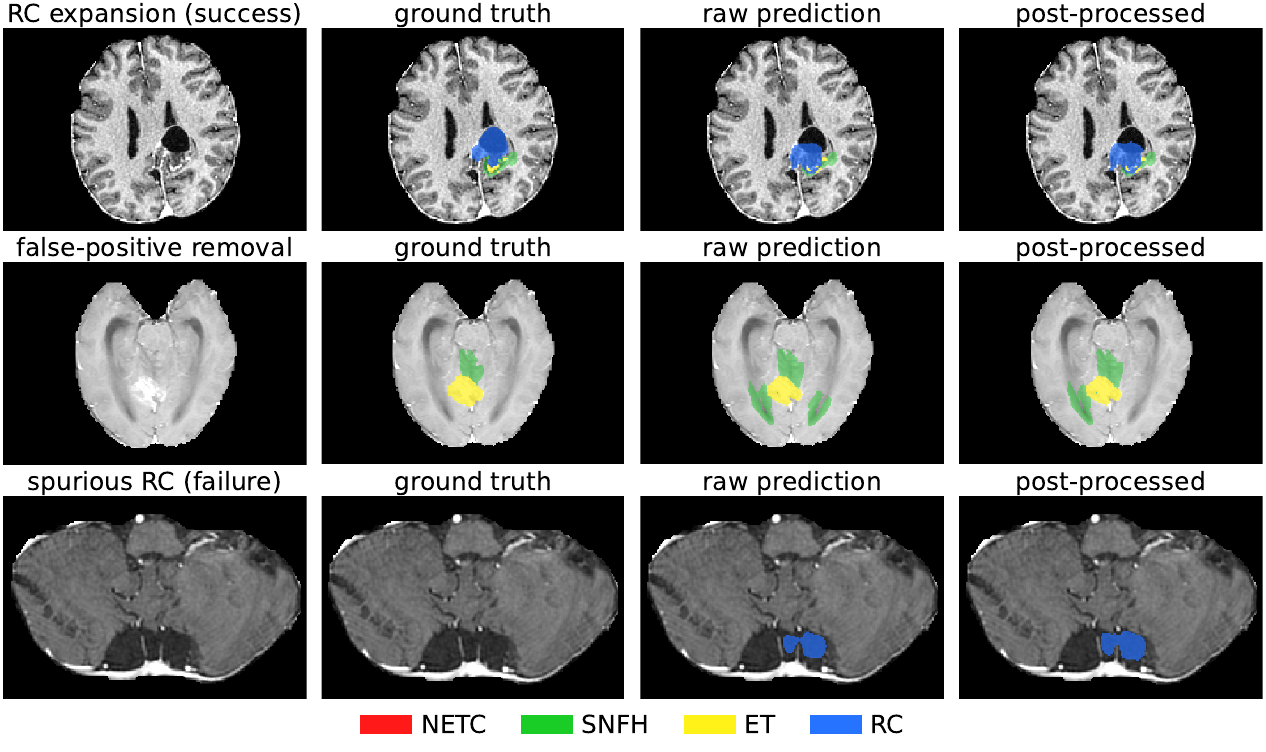}
\caption{Raw-versus-post-processed comparison on three fold-0 held-out cases
(columns, left to right: T1c input, ground truth, raw ensemble, post-processed with
\texttt{RuleClean}~+~\texttt{BoundaryExpand}). The first column shows the T1c input,
titled with the phenomenon that row illustrates. Top: \texttt{BoundaryExpand} grows
an under-segmented RC cavity toward the ground-truth extent. Middle:
\texttt{RuleClean} removes a spurious sub-threshold component. Bottom: a failure
case --- the raw model predicts a spurious RC on a baseline scan (no GT cavity)
that the cascade does not suppress. Overlays: NETC (red), SNFH (green), ET
(yellow), RC (blue).}
\label{fig:qualitative}
\end{figure}

\begin{table}[t]
\centering
\small
\setlength{\tabcolsep}{4.5pt}
\caption{Post-processing ablation on the full five-fold OOF ($1{,}296$ cases;
official BraTS-METS/Panoptica evaluator, per-region means over
ground-truth-present cases). Stages are cumulative: \texttt{RuleClean}, then
\texttt{BoundaryExpand} (\emph{recommended} cascade), then \texttt{brainF70}
(\emph{submitted}; a negative result, \Cref{sec:failed}). \texttt{BoundaryExpand}
and \texttt{brainF70} modify only RC, so ET\slash TC\slash WT are unchanged after
\texttt{RuleClean}. avg4 is the mean over the four regions;
$\uparrow$/$\downarrow$: higher/lower is better.}
\label{tab:ablation}
\begin{tabular}{@{}ll ccccc@{}}
\toprule
Metric & Configuration & ET & TC & WT & RC & avg4 \\ \midrule
\multirow{4}{*}{DSC $\uparrow$}
 & Raw                        & 0.6163 & 0.6395 & 0.5968 & 0.3667 & 0.5548 \\
 & + \texttt{RuleClean}       & 0.6643 & 0.6799 & 0.6400 & 0.4911 & 0.6188 \\
 & + \texttt{BoundaryExpand}  & 0.6643 & 0.6799 & 0.6400 & 0.4991 & 0.6209 \\
 & + \texttt{brainF70}        & 0.6643 & 0.6799 & 0.6400 & 0.4971 & 0.6203 \\ \addlinespace
\multirow{4}{*}{NSD $\uparrow$}
 & Raw                        & 0.690 & 0.700 & 0.618 & 0.296 & 0.576 \\
 & + \texttt{RuleClean}       & 0.739 & 0.739 & 0.655 & 0.381 & 0.629 \\
 & + \texttt{BoundaryExpand}  & 0.739 & 0.739 & 0.655 & 0.377 & 0.628 \\
 & + \texttt{brainF70}        & 0.739 & 0.739 & 0.655 & 0.374 & 0.627 \\ \addlinespace
\multirow{4}{*}{HD95\,(mm) $\downarrow$}
 & Raw                        & 90.7 & 86.4 & 94.7 & 170.9 & 110.7 \\
 & + \texttt{RuleClean}       & 72.4 & 71.8 & 79.6 & 112.3 & 84.0 \\
 & + \texttt{BoundaryExpand}  & 72.4 & 71.8 & 79.6 & 109.5 & 83.3 \\
 & + \texttt{brainF70}        & 72.4 & 71.8 & 79.6 & 113.2 & 84.2 \\ \addlinespace
\multirow{4}{*}{Large-inst.\ F1 $\uparrow$}
 & Raw                        & 0.752 & 0.761 & 0.765 & 0.070 & 0.587 \\
 & + \texttt{RuleClean}       & 0.792 & 0.794 & 0.799 & 0.087 & 0.618 \\
 & + \texttt{BoundaryExpand}  & 0.792 & 0.794 & 0.799 & 0.089 & 0.619 \\
 & + \texttt{brainF70}        & 0.792 & 0.794 & 0.799 & 0.088 & 0.618 \\ \addlinespace
\multirow{4}{*}{Small-inst.\ F1 $\uparrow$}
 & Raw                        & 0.360 & 0.368 & 0.302 & 0.021 & 0.263 \\
 & + \texttt{RuleClean}       & 0.277 & 0.279 & 0.223 & 0.017 & 0.199 \\
 & + \texttt{BoundaryExpand}  & 0.277 & 0.279 & 0.223 & 0.017 & 0.199 \\
 & + \texttt{brainF70}        & 0.277 & 0.279 & 0.223 & 0.017 & 0.199 \\ \bottomrule
\end{tabular}
\end{table}

Three findings survive this larger audit. First, \texttt{RuleClean} is the
dominant and unambiguous contributor: it increases avg4 by $+0.0640$ and RC by
$+0.1244$, with a positive per-fold avg4 increment on all five folds (range
$+0.052$ to $+0.082$). Second, \texttt{BoundaryExpand} is a small but consistent
RC operator: $+0.0080$ RC ($+0.0021$ avg4), non-negative on every fold and with
no change to the ET\slash TC\slash WT channels. Third, \texttt{brainF70} does
\emph{not} replicate its leaderboard behaviour: overall it changes avg4 by
$-0.0006$ and RC by $-0.0020$, is triggered on only $6$ of the $1{,}296$ cases,
and reverses sign across folds --- removing a true false positive on fold~0, but
deleting a genuine resection cavity on fold~2. For reference, on the leaderboard
the \texttt{RuleClean}$\to$full increment is $+0.0164$ RC ($0.5323\to0.5487$); our
audit attributes the durable part of this to \texttt{BoundaryExpand} and the
remainder to leaderboard noise from a single \texttt{brainF70} deletion. Because this net effect
lies within fold-to-fold noise rather than a reliable regression, we exclude
\texttt{brainF70} from the recommended cascade and report it as a negative
result (\Cref{sec:failed}); it was present in the submitted Docker container
(\Cref{tab:main}) only to reproduce the leaderboard score, and is in any case
inert on all but a small number of cases.
Beyond lesion-wise DSC, \Cref{tab:ablation} also reports the additional
official metrics requested for a complete assessment (per-region lesion-wise
NSD, HD95, and large- and small-instance F1) for the raw ensemble and each
cumulative stage. \texttt{RuleClean} improves NSD, HD95, and large-instance F1
in every region and roughly halves HD95, confirming that it sharpens both
overlap and boundary fidelity; it does lower small-instance F1 (e.g.\ ET
$0.360\to0.277$), the direct and intended consequence of deleting
sub-$10\,\text{mm}^{3}$ components, a cost outweighed on the size-dominant
lesion-wise and large-instance metrics. \texttt{BoundaryExpand} improves RC HD95
($112.3\to109.5$~mm) alongside its RC DSC gain, so the overlap improvement does
not come at the expense of surface fidelity.

\Cref{fig:qualitative} shows the raw versus post-processed output on three
fold-0 cases (a success, a false-positive removal, and a failure).

\section{Failed attempts}
\label{sec:failed}

On the training side, a Dice$+$top-$k$ CE loss ($k=10\%$) improved ET and TC by
under a point each but degraded RC by more than two, and a routing variant
feeding only the RC channel from the top-$k$ model reproduced the failure. A
MedNeXt-L backbone added as a sixth ensemble member degraded \texttt{avg4} under
both an equal $1/6$ and a $0.7/0.3$ ResEnc-L/MedNeXt weighting. Two Focal-Tversky
variants also failed: $\gamma=0.75$ collapsed ET and RC Dice to zero within a
hundred epochs, and a milder $\gamma$ ran to epoch~613 without exceeding the
DC$+$CE baseline. We were thus unable to compensate RC's extreme rarity through
the loss alone at this dataset scale.

On the post-processing side, two aggressive volume filters (removing RC
components below $100$ or $300$ voxels) degraded RC by up to seven points, and a
five-iteration ungated ring dilation produced the study's largest RC regression
by extending into skull and CSF; these bracket the conservative, gated $i=3$
\texttt{BoundaryExpand}. \texttt{brainF70} (Stage~3, deleting RC components with
brain fraction $<0.70$) helped one leaderboard case but was net-negative and
sign-reversing across the five OOF folds (\Cref{sec:ablation}), so it is excluded
from the recommended cascade. Lowering the RC softmax threshold to $0.40$ traded
precision for recall at a net loss; an SNFH $<10\,\text{mm}^{3}$ filter was
neutral and dropped; four region-swap fusion variants gave nothing once an axial
I/O bug was fixed; and a $0.25$ sliding-window sweep (75\% overlap, three
weightings) was consistently worse than $0.5$ (\texttt{avg4}
$0.6795$--$0.6796$ vs $0.6823$), which we attribute to over-averaging at patch
boundaries.

\section{Discussion}
\label{sec:discussion}

\noindent\textbf{Post-processing carries more weight under LW-DSC than under global Dice, but only in one direction.} Prior BraTS write-ups report that
post-processing beyond light small-blob filtering is nearly
neutral~\cite{krikorian2025multiarch,monaidints2025brats}; under voxel-wise
(global-volume) Dice this is expected, as a single small component is negligible
as a voxel fraction. Lesion-wise DSC changes this: it matches components
case-by-case and averages Dice \emph{per lesion}, so that same component becomes
an entire term in the mean, creating a sharp asymmetry. For a $200$-voxel
ground-truth RC lesion, growing a partial prediction from $80$ to $160$ voxels
lifts that lesion's Dice from $0.57$ to $0.89$, whereas $40$ spurious voxels on
an otherwise perfect match cost only $1.0\!\to\!0.91$ --- yet deleting the
component outright scores it $0$. Recall-recovering operations therefore sit on
the safe side of this asymmetry, while deletion is the single most expensive event.

\smallskip
\noindent Our experiments trace out exactly this safe band. Confidence-gated
boundary growth (\texttt{BoundaryExpand}) recovers RC recall with a
sign-consistent gain at no cost to the other channels, whereas both departures
from the band (over-deletion by volume filters and ungated over-expansion)
regress RC (\Cref{sec:ablation,sec:failed}). The practical rule for LW-DSC
post-processing is thus narrow but robust: recover recall conservatively under a
probability gate, and never delete a component unless it is confidently a false
positive.

\smallskip
\noindent\textbf{Remaining RC gap.} Our final RC LW-DSC of $0.549$ remains well
below the top validation submissions. We conjecture the highest-yielding RC
levers lie on the training side (foreground oversampling, a connected-component
loss on RC) and detection side (an RC-specialised head) rather than in
post-processing; a dedicated RC pathway is our first-priority extension.

\begin{credits}
\subsubsection{\ackname}
We thank the BraTS 2026 organisers for curating the challenge datasets, and the
Synapse team for hosting the evaluation infrastructure.

\subsubsection{\discintname}
The authors have no competing interests to declare.
\end{credits}

\bibliographystyle{splncs04}
\bibliography{refs}

\end{document}